\documentclass[letterpaper, 10 pt, conference]{ieeeconf}  

\IEEEoverridecommandlockouts                              

\usepackage{graphics} 
\usepackage{epsfig} 
\usepackage{amsmath} 
\usepackage{amssymb}  
\usepackage{algorithm}
\usepackage{algpseudocode}
\usepackage{float}
\usepackage{lipsum}  
\usepackage[dvipsnames]{xcolor}
\usepackage{caption}
\usepackage{subcaption}
\usepackage{comment}
\usepackage{multirow}
\usepackage[hidelinks]{hyperref}
\title{\LARGE \bf
Leveraging Pretrained Latent Representations for Few-Shot Imitation Learning on a Dexterous Robotic Hand
}

\author{Davide Liconti$^{1}$, Yasunori Toshimitsu$^{1,2}$ and Robert Katzschmann$^{1}$ %
\thanks{$^{1}$Soft Robotics Lab, IRIS, D-MAVT, ETH Zurich, Switzerland
        {\tt\footnotesize \{dliconti, ytoshimitsu, rkk\}@ethz.ch}}%
\thanks{$^{2}$Max Plank ETH Center for Learning Systems}%
}

\begin{document}

\maketitle
\thispagestyle{empty}
\pagestyle{empty}

\begin{abstract}
In the context of imitation learning applied to dexterous robotic hands, the high complexity of the systems makes learning complex manipulation tasks challenging. 
However, the numerous datasets depicting human hands in various different tasks could provide us with better knowledge regarding human hand motion.
We propose a method to leverage multiple large-scale task-agnostic datasets to obtain latent representations that effectively encode motion subtrajectories that we included in a transformer-based behavior cloning method.
Our results demonstrate that employing latent representations yields enhanced performance compared to conventional behavior cloning methods, particularly regarding resilience to errors and noise in perception and proprioception.
Furthermore, the proposed approach solely relies on human demonstrations, eliminating the need for teleoperation and, therefore, accelerating the data acquisition process. Accurate inverse kinematics for fingertip retargeting ensures precise transfer from human hand data to the robot, facilitating effective learning and deployment of manipulation policies.
Finally, the trained policies have been successfully transferred to a real-world 23Dof robotic system.
\end{abstract}

\section{INTRODUCTION}

\subsection{Motivation}   

Dexterous manipulation constitutes an intriguing domain within the field of robotics, primarily due to its capacity to facilitate the execution of a wide spectrum of tasks that can benefit from the coordinated movement of multiple fingers. Additionally, a dexterous manipulator with many degrees of freedom mimicking the human hand can be used as a universal manipulator.
However, the high number of degrees of freedom required to achieve dexterity also complicates the application of conventional control approaches.

Reinforcement learning (RL), which has been widely used for dexterous manipulation research, offers a possible solution. Recent advancements in parallelized GPU-based simulators have made it feasible to train dexterous policies within a few hours.
However, there are some constraints and challenges associated with RL, such as the necessity of creating a dedicated simulation environment and reward for each unique policy and task. Additionally, overcoming the "sim-to-real" gap between the simulated physics and the real world can require considerable effort.
\begin{figure}
    \centering
    \includegraphics[width=\columnwidth]{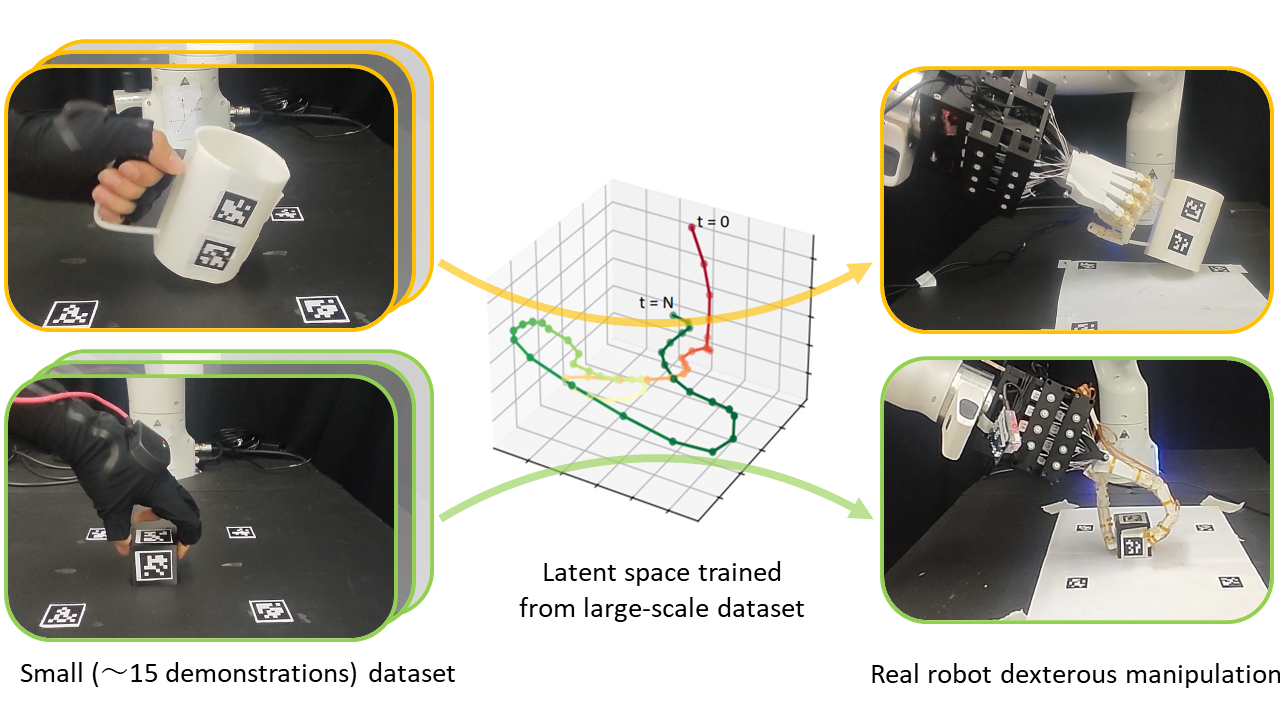}
    \caption{Compared with traditional behavior cloning approaches, this work proposes a few-shot end-to-end pipeline that uses pre-trained latent representations learned from multiple large-scale task-agnostic datasets. This latent space effectively encodes robot actions, giving benefits regarding robustness and stability of the learned policies. Moreover, manipulation demonstrations are acquired without teleoperation, and then deployed on a real dexterous robotic system. }
    \label{fig:teaser}
\end{figure}

Another approach is to learn behaviors from demonstrations in a supervised learning fashion, referred to as behavior cloning (BC). As the policy can learn from human demonstrations rather than exploring the environment on its own (like in RL), BC could be more efficient and more capable of replicating human-like behavior.
\begin{figure*}
  \includegraphics[width=0.99\textwidth]{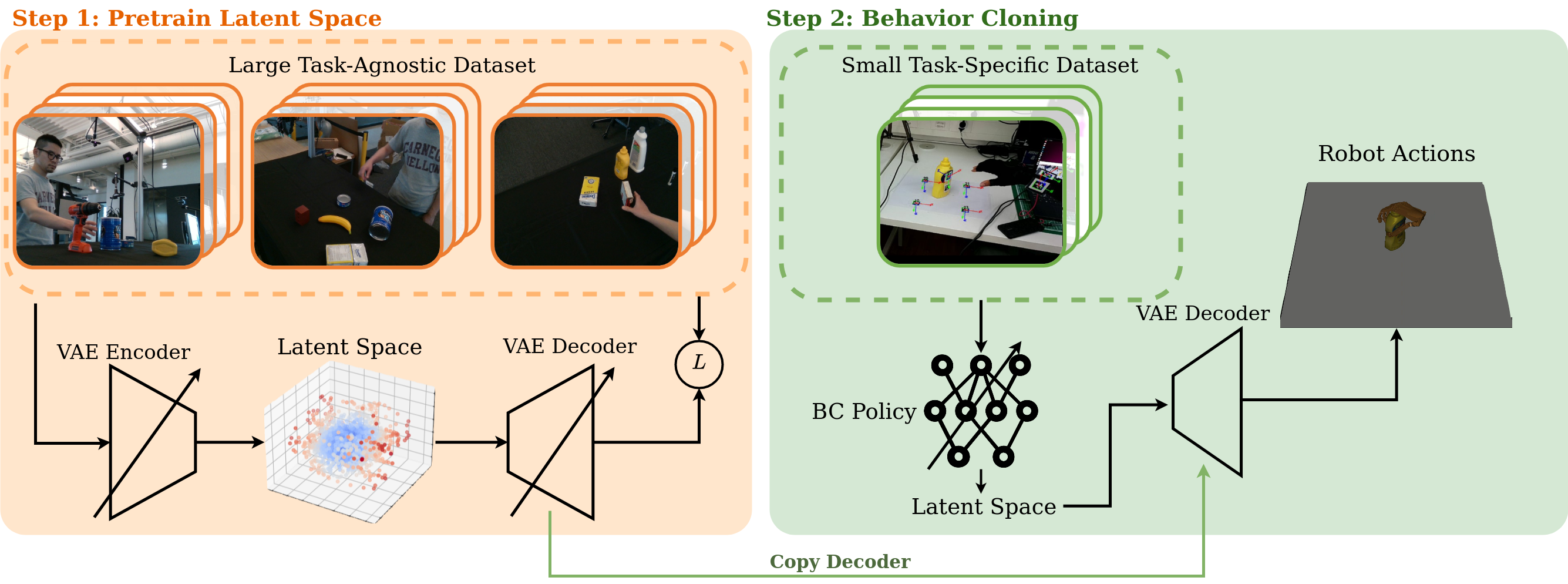}
  \caption{
  Our imitation learning method leverages latent representations of valid human hand motion, pre-strained on large-scale datasets. \textit{Step 1:} We train a reconstruction-based VAE to learn a latent space representation of hand motion by leveraging multiple large task-agnostic datasets, reducing the dimension needed to encode hand subtrajectories. \textit{Step 2:} We collect a small dataset of recorded demonstrations with a data acquisition pipelines that tracks object and poses, and train a behavior policy that outputs latent representations of the hand trajectories, which are then decoded with the pretrained decoder.}
    \label{fig:method_general}
    \end{figure*}
 The prevailing method for acquiring demonstrations involves teleoperation, during which the robot is guided to execute specific tasks while concurrently gathering various data inputs that are used to train policies. While this approach can indeed prove effective and resilient, data collection through teleoperation requires robot availability and a specific hardware setup for an extended duration, which poses a significant issue in hypothetical industrial scenarios. Consequently, there exists a pressing need to develop an alternative method for imitation learning that does not rely on costly teleoperation-based data, and directly uses human demonstrations. With robotic platforms like anthropomorphic hands, the biomimetic nature of the system can be used to establish a consistent mapping from human hand actions into the robot state space, making it feasible to use human demonstrations directly.

However, a critical challenge with BC techniques is the substantial quantity of demonstrations required for robust results. Contemporary state-of-the-art approaches have relied upon a considerable volume of input data, often numbering in the hundreds, even for relatively straightforward tasks. This demand for such large data collection for every new task is very time and effort-inefficient. Therefore, there is a need for methods that can mitigate the necessity for an extensive number of demonstrations while simultaneously preserving the robustness of the learned policies.

\subsection{Approach and Contributions}
This work proposes to develop a task-agnostic method to learn dexterous manipulation policies from human demonstration data, with a specific emphasis on preserving the robustness of the policies in a low data regime.\\
By eliminating the need for teleoperation data, it is possible to accelerate the data acquisition procedure, making it accessible to non-expert users and eliminating the need for robot hardware availability during the process. \\
Additionally, building upon the observation that there are already many datasets of human hands doing various tasks publicly available, this work suggests a way to leverage these datasets to give the robot a better knowledge of how human hands move, therefore letting the policy learn efficiently even from a relatively small task-specific dataset.

The contributions of this work are:
\begin{itemize}
    \item Propose a pipeline to acquire human demonstrations of manipulation tasks and retarget them into the robot state space;
    \item Get latent representation of hand motion by leveraging multiple large scale datasets;
    \item Use the pretrained representation to increase robustness of behavior cloning methods, reducing the number of required demonstrations for an effective translation of behaviors on real-world dexterous robots.
\end{itemize}

\section{RELATED WORKS}

\subsection{Learning-based control for Dexterous Manipulation} 
A substantial body of research has been dedicated to learning-based control techniques for dexterous manipulation. Several works use pure RL approaches, effectively applying a closed-loop RL policy for a biomimetic robot hand \cite{openai2019learning}. This approach has also been applied to the Faive Hand~\cite{toshimitsu2023getting}, the robotic hand used in this work.

In addition to these efforts, multiple approaches have been proposed to incorporate demonstrations into learning-based control frameworks. These methods involve integrating demonstrations into online RL approaches or employing simple supervised learning techniques like behavior cloning (BC). Notably, \cite{rajeswaran2018learning} introduced DAPG (Demo Augmented Policy Gradient), one of the most commonly used algorithms to bootstrap policy gradient RL algorithms with demonstrations. Furthermore, \cite{radosavovic2021stateonly} and \cite{ho2016generative} are among other examples showcasing the integration of demonstrations into online RL setups. \\
On the other hand, with BC the agent policy is trained to replicate a set of given demonstrations directly. BC from teleoperations is showing incredible capabilities to make robot learn very diverse tasks, with a high-level of precision, as demonstrated by \cite{shridhar2022perceiveractor},\cite{zhao2023learning},  \cite{fu2024mobile}, \cite{brohan2023rt1}, \cite{brohan2023rt2} and \cite{chi2023diffusion}.
In the context of dexterous manipulation with antropomorphic hands, in the literature there is a variety of approaches both on the input data and the imitation learning method. \cite{arunachalam2022dexterous} uses real-world teleoperation data, \cite{qin2023hand} uses teloperations in simulation evironments, making faster and cheaper to acquire demonstrations, \cite{wang2024cyberdemo} augments teleoperation data with simulated human demonstrations, while \cite{qin2022dexmv} or \cite{wang2024dexcap} directly uses the human demonstrations as source data. An alternative approach is also to use the robot gripper directly moved by a human \cite{chi2024universal}.

\subsection{Representation Learning for Imitation}
One approach to increase the robustness over long-horizon tasks is to learn a latent space representation of the target motion (i.e. skills).  \cite{pertsch2021demonstrationguided}, \cite{ajay2021opal} and \cite{wang2023mimicplay} use skill-based frameworks to integrate online or offline RL setups. By performing learning on a skill space extracted from a large task-agnostic dataset we could also mitigate the need for many downstream demonstrations by exploiting that previous knowledge. \cite{nasiriany2022learning} uses the learned latent space also to retrieve data from larger task-agnostic datasets using similarity functions. \cite{ye2023learning} focuses on finding a continuous grasping function for a dexterous gripper by leveraging diverse human demonstrations, using a Conditional Variational Autoencoder to find this implicit grasping representation. Finally, \cite{gehring2023leveraging} shows how to exploit latent priors for different applications in RL control.

\section{HARDWARE SETUP}
\label{hardware}
Our experimental setup consists of a 7Dof Franka Emika 3 arm, on which we attached the anthropomorphic \textit{Faive} hand (\cite{faive,toshimitsu2023getting}). The hand has 16 Dof -- 4 in the thumb and 3 on each of the other fingers. The hand is tendon-driven and actuated by 16 servo motors and uses rolling contact joints between the phalanges.

\section{METHOD}
\subsection{Overview}
The final goal of our work is to replicate human behaviors on a dexterous hand without teleoperation data, using a relatively small task-specific dataset. To address this challenge we tried to perform behavior cloning within a learned lower dimensional latent space.
 We first use a large offline dataset $\mathcal{D}_{\text{prior}}$ from task-agnostic interactions to learn a latent space $\mathcal{Z} \in \mathbb{R}^D$ representation of fixed-length short-horizon sub-trajectories $\tau = \{a_0, a_1, \ldots, a_N\}$ with a reconstruction-based Variational Autoencoder. Open-source manipulation datasets can be used for $\mathcal{D}_{\text{prior}}$. In the next step, we prepare a small offline dataset $\mathcal{D}_{\text{task}}$ specifically collected for the target task and train a policy $\pi$ that predicts latent space representations $\mathbf{z} \in \mathcal{Z}$ based on the current observations, which can be decoded into action sequences via the pretrained decoder. A summary of the proposed method is shown in Fig. \ref{fig:method_general}.
Our work also presents a perception pipeline to acquire downstream tasks human demonstrations of $\mathcal{D}_{task}$, explained in the next section.

\begin{figure}
    \centering
    \includegraphics[width=0.48\textwidth]{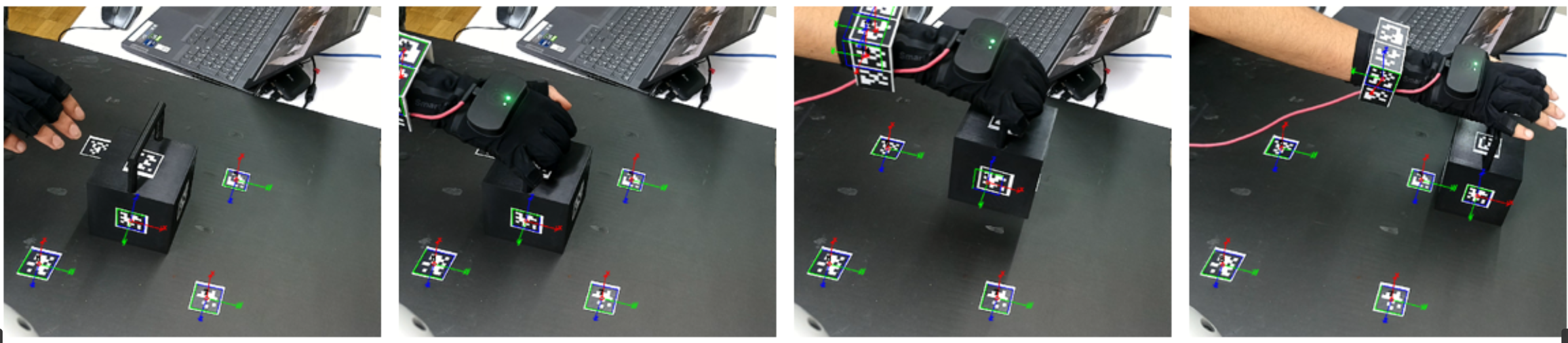}
    \caption{Data collection for the small task-specific demonstrations. A motion capture glove captures the finger motions, and visual markers are used for tracking the wrist and object poses. With this pipeline the data acquisition can be considerably sped up with respect to teleoperation-based methods.}
    \label{fig:data_acquisition}
\end{figure}

\begin{figure}
    \centering
    \includegraphics[width=0.48\textwidth]{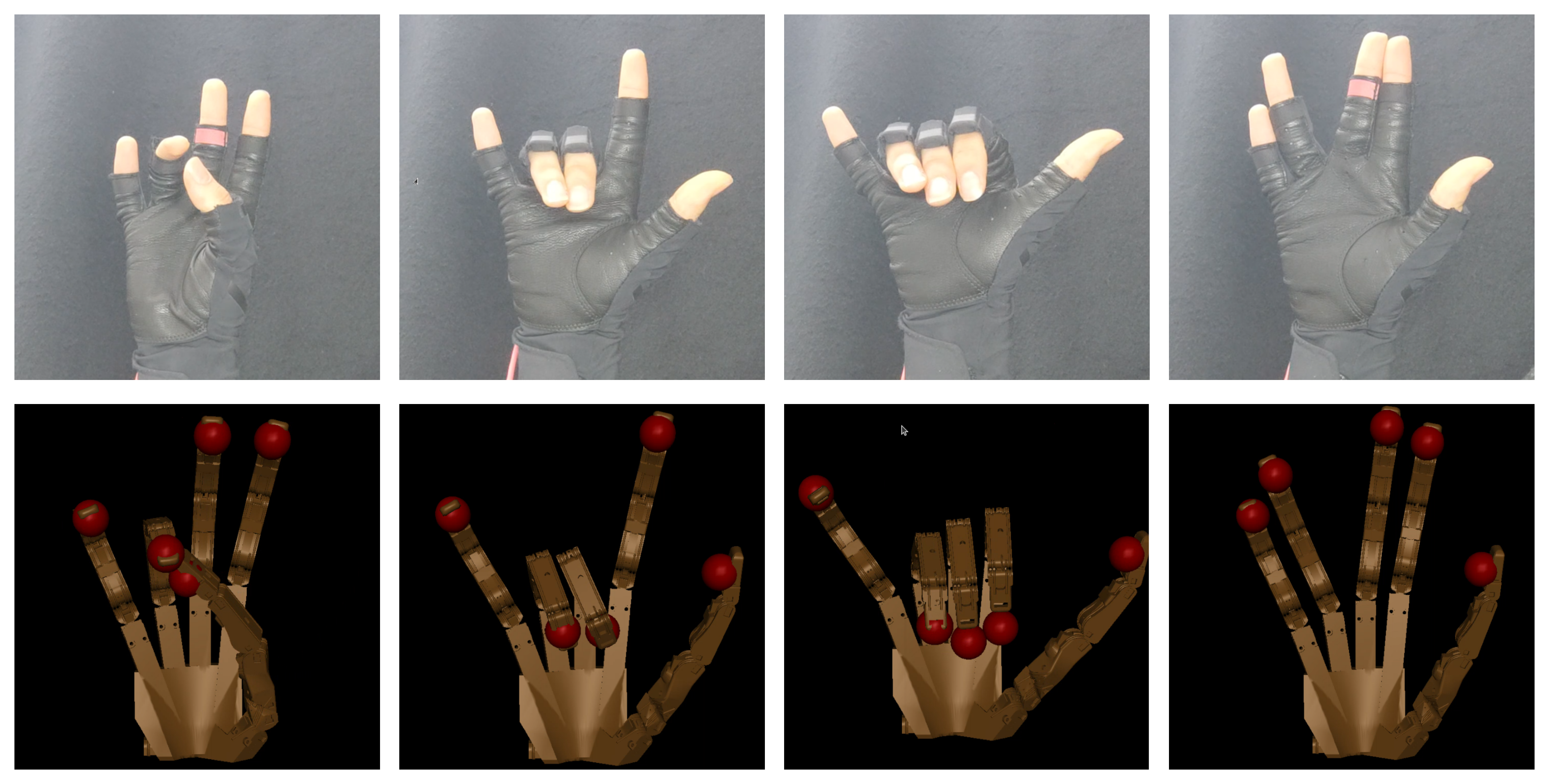}
    \caption{Visual results of IK retargeting. The red dots represent the 3D positions of the fingertips coming from the motion capture glove. Due to the 16Dof of the hand, we can mimic very extreme poses, including pure finger abduction.}
    \label{fig:retargeting}
\end{figure}

\subsection{Data Acquisition}
To gather human demonstrations efficiently, the perception pipeline should prioritize speed and ease of use. For instance, precise camera calibration should not be a requirement, as our goal is to maximize the speed of the data acquisition process. This approach should ensure that even non-trained users can proficiently collect the demonstrations.
A demonstration consists of a sequence of states, each composed of the full human hand pose and the object pose for tasks involving interaction with an object. 
In this study, data reliability is heavily emphasized, given that inaccurate recordings inevitably lead to erroneous robot behaviors. Hand poses are acquired using a motion capture glove \cite{rokoko}, which employs hybrid IMU and EMF (Electromagnetic Fields) sensor fusion. Wrist and object poses are captured using Apriltag markers \cite{apriltags}, expressed in a world frame identified by tags placed on the table. A Kalman filter for position and velocity is employed to smooth and interpolate the results. The pipeline operates in real-time at approximately 25 fps.
This method accelerates the data acquisition process substantially. In particular, the average total acquisition time for one demonstration is reduced by about 75\% if compared with teleoperation using the same robotic setup.
It's worth noting that current computer vision technologies offer the possibility of a vision-based data acquisition pipeline without the need for tags or a motion capture glove. Several studies have focused on estimating 3D hand poses from RGB \cite{zimmermann2017learning}, \cite{rong2020frankmocap}, \cite{liu2021semisupervised} or RGB-D images \cite{rezaei2022trihornnet}. These models often utilize a parametric hand model \cite{MANO} (MANO) identifying 21 key points representing hand joints.
Similarly, numerous methods are available for object tracking in real-world scenarios, with state-of-the-art approaches not requiring prior knowledge of the object's geometry \cite{wen2023bundlesdf}.
However, since the primary focus of this work is not the data collection method, we have opted for simpler solutions to obtain a similar set of data.
\subsection{Motion Retargeting}
In the specific context of dexterous manipulation, already some attempts have been made to retarget poses from human hands to robotic hardware. Antotsiou et al. combined inverse kinematics and PSO with a task objective optimization \cite{antotsiou2018taskoriented}, while Sivakumar et al. and Handa et al. define keyvectors on both the human and the robot hand and minimizes an energy function to kinematically retarget the hand poses to the robot space  \cite{sivakumar2022robotic, handa2019dexpilot}.
However, such online optimization methods, although they can effectively mimic a hand pose, sometimes do not recover the exact 3D position of the fingertips, which is crucial to have effective grasping or manipulation behaviors. The robot hand that we used has at least 3 Dof on each finger, making it possible to solve the inverse kinematics (IK) independently for each finger. Thus, it is possible to have an exact mapping of the fingertip positions.\\
By numerically solving the inverse kinematics, the retargeter achieved a speed of 80Hz, making it suitable for real-time teleoperation scenarios as well. Visual results are provided in Fig. \ref{fig:retargeting}.

\subsection{Latent Space Training}
\begin{figure}[h]
    \centering
    \includegraphics[width=0.49\textwidth]{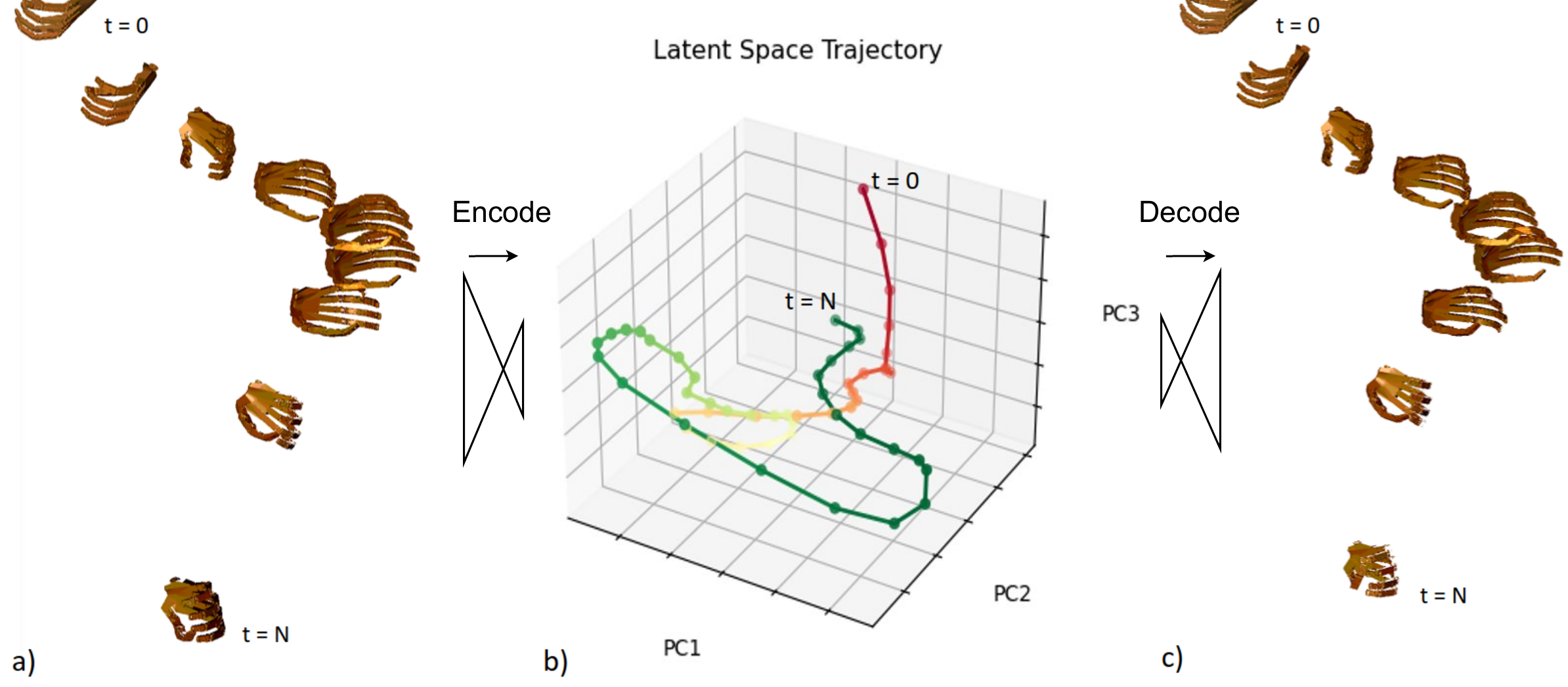}
    \caption{Analysis and reconstruction of input demonstrations, where the input is retargeted into the robot state space, projected into a latent space for smooth trajectory analysis thanks to KL divergence loss, and finally reconstructed with high fidelity to the original demonstration. \\a) Retargeted input demonstration, sampled every 7 frames. \\ b) Latent space projection (PCA) of the subtrajectories of the input demonstration, with a sliding window of size 1. We can notice that the latent space trajectory is smooth and continuous, due to the KL divergence loss term. \\ c) Reconstruction of the input demonstration, obtained by taking all the frames of the first latent space representation, and the last frame of all the subsequent decoded subtrajectories. We can notice a good reconstruction by comparing this visualization with the input demonstration.}
    \label{fig:reconstruction_latent}
\end{figure}

\begin{figure}[h]
    \centering
    \includegraphics[width=0.25\textwidth]{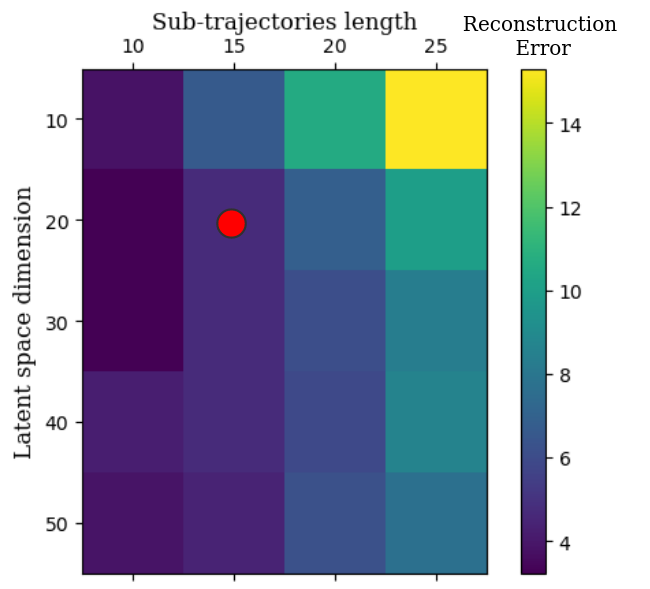}
    \caption{Comparison study to determine the hyperparameters for the reconstruction based VAE. the value of the grid represents the reconstruction error of the sequences
    We aim for a high subtrajectory length, so that longer horizon motions can be embedded, and low latent dimension, to express the motion with as few parameters as possible, while having a good reconstruction. The "sweet spot" was a latent dimension of 20 (5.3\% of the original dimension) and a subtrajectory length of 15 (0.5s motion), marked in red. }
    \label{fig:ablation}
\end{figure}
\begin{figure}[h]
      \centering
    \includegraphics[width=0.49\textwidth]{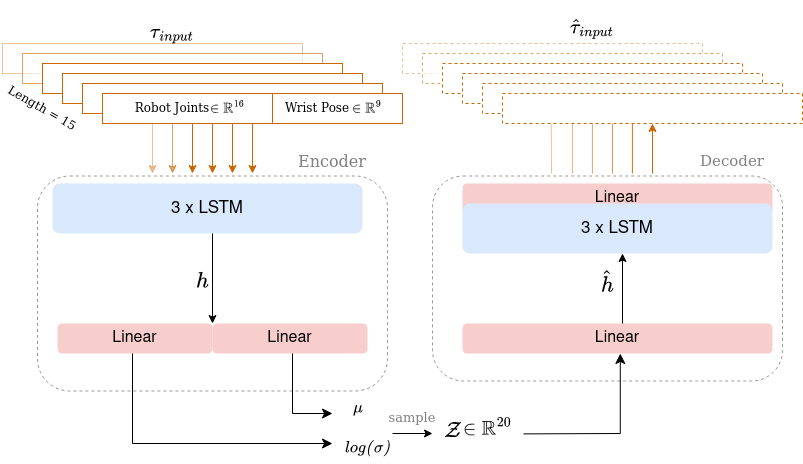}
    \caption{The VAE consists of an LSTM-based encoder-decoder architecture, with linear layers to get to the wanted latent dimension. Between the LSTM modules, tanh activation functions  are used.}
    \label{fig:vae_architecture}
\end{figure}
The first step of our method consists of obtaining a meaningful latent representation of hand motion.
As the dataset we take the demonstrations from a set of multi-task human hand motion datasets. We merged the DexYCB  \cite{chao2021dexycb}, ARCTIC \cite{fan2023arctic} and GRAB \cite{Taheri_2020} datasets, that contain sequences of human hands poses, while performing a broad set of tasks. The combination of the three datasets provides almost 3M annotated frames. All the sequences from the datasets have been sampled at 30fps.
We then retargeted all the demonstrations into the robot state space with the IK approach described in the previous section, allowing the motion to be in the robot state space. 

The input data is composed of the joints angles of the robot hand $q_{robot} \in \mathbb{R}^{16}$, the wrist position $q_{w, pos} \in \mathbb{R}^3 $ and a 6D representation of the wrist orientation given by the concatenation of the first 2 columns of the rotation matrix $q_{w, rot} \in  \mathbb{R}^6 $. Given this representation the rotation matrix can always be recovered from the cross product between the two vectors. This orientation representation was chosen since, in a sequence of rotations, lower dimensional representations like Euler angles or quaternions are not guaranteed to be continuous \cite{zhou2020continuity}, making it more difficult to be processed by a neural network.

For generating latent space representations of hand motion, we sample short-horizon subtrajectories from the retargeted demonstrations, denoted as $\tau = \{a_0, a_1, \ldots, a_N\} \in \mathbb{R}^{N \times 25}$, with a length of $N$. The subtrajectory length and the latent space dimension are hyperparameters on which we perform a comparison study in figure \ref{fig:ablation}, and trained a Variational Autoencoder (VAE) to reconstruct these sequences.

To account for the sequential nature of the data, we employed a Long Short-Term Memory (LSTM) encoder $q_\phi$ to encode the subtrajectories into a hidden space representation $h \in \mathbb{R}^H$. Subsequently, a Multilayer Perceptron (MLP) further reduces it into a Gaussian latent distribution $z \in \mathbb{R}^D$. The decoding component consists of an MLP that reconstructs the hidden representation $\hat{h}$ and an LSTM decoder $p_{\psi}$ that produces the reconstructed subtrajectory $\hat{\tau}$.

The loss objective is defined as follows:
\begin{equation}
    \mathcal{L}_{VAE} = \parallel \tau - \hat{\tau} \parallel _2^2 + \parallel h - \hat{h} \parallel_2^2 + \gamma \mathcal{D}_{KL}\left(q_{\phi}(z|\tau)\parallel\mathcal{N}(0,1)\right)
\end{equation}

Here, $\gamma$ controls the effect of the KL divergence term, encouraging the latent space representations to be predictable and smooth by minimizing the difference between the latent space distribution $q_{\phi}(z|\tau)$ and the standard normal distribution. An outline of the VAE architecture can be seen in fig \ref{fig:vae_architecture}.

\subsection{Behavior Cloning Policy}
\subsubsection{Policy Training}
For the downstream policy learning task, we train a policy, denoted as $\pi$, designed to take a sequence of the past $L$ robot states $\tau_{input} = {s_i, s_{i+1},..., s_{i+L}}$ as input. This policy predicts the latent space representation $z$ of the sequence shifted by $n$ frames. Then the latent space representation is decoded with the pretrained decoder $p_\psi$ into the action sequence $\tau_{output} = \{a_{i+n}, a_{i+n+1},..., a_{i+n+N}$\}, where $N$ is the sub-trajectory length used to train the VAE. 

For tasks that require an interaction with an object(e.g. grasping, pick-and-place, etc...), we augment the input data also with the object pose $s_{obj} \in \mathbb{R}^7$, composed by the position and quaternion of the object.

The policy network is an adapted version of the Action Chunking with Transformers (ACT) architecture from \cite{zhao2023learning}. 
The loss is computed minimizing the difference between the predicted and the original shifted sequences: 
\begin{equation}
    \mathcal{L}_{BC} = \parallel \tau_{output} - p_{\psi}(\pi(\tau_{input}, s_{obj})) \parallel _2^2
\end{equation}

\subsubsection{Data Pre-Processing}

\begin{figure}
    \centering
        \begin{subfigure}[b]{0.21\textwidth}
        \centering
        \includegraphics[width=\textwidth]{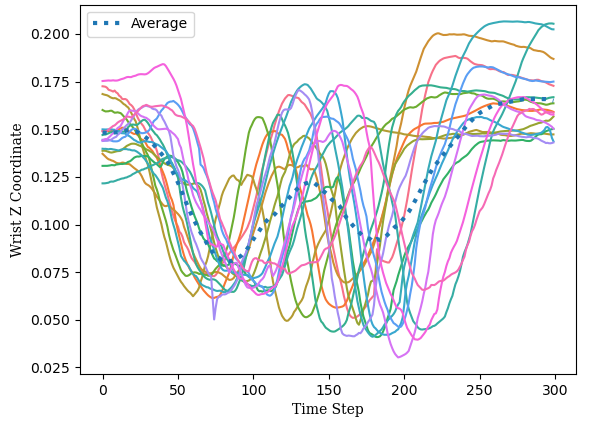}
        \caption{}
        \label{fig:data_post_processing_left}
    \end{subfigure}
    \begin{subfigure}[b]{0.23\textwidth}
        \centering
        \includegraphics[width=\textwidth]{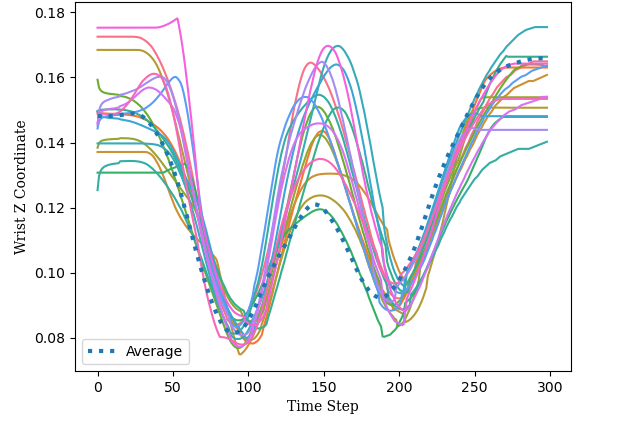}
        \caption{}
        \label{fig:data_post_processing_right}
    \end{subfigure}
    \label{fig:data_post_processing}
    \caption{\textbf{a):} Raw data from the demonstration. \textbf{b):} after postprocessing the data with a low pass filter and dynamic time warping to temporally align the data.}
\end{figure}

We preprocess the trajectories by filtering them with a moving average low-pass filter to reduce the noise coming mainly from jittery tag detections, and then we apply the Dynamic Time Warping (DTW) algorithm to temporally align the trajectories, such that the phase and the speed of the movements are aligned (Fig. \ref{fig:data_post_processing_right}).

\subsubsection{Inference}

\begin{figure}
    \centering
    \includegraphics[width=0.4\textwidth]{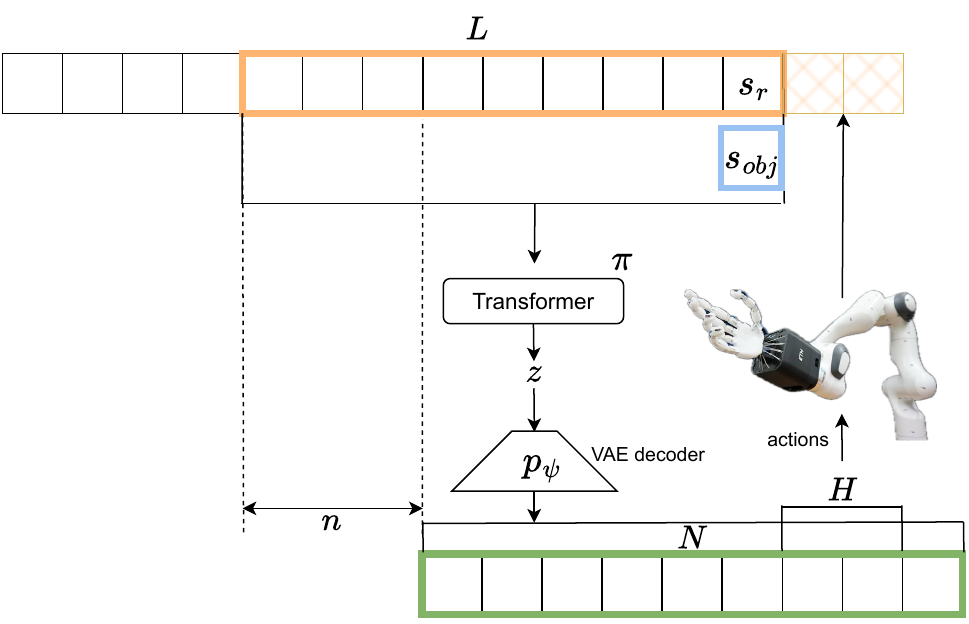}
        \caption{The relation of the data input to and output from the policy. It receives the past $L$ states, converts it to a vector $z$ in the latent space, and the pretrained decoder is used to convert it to a sequence of $N$ robot actions. The next $H$ actions are executed on the robot, after which the process is repeated.}
    \label{fig:inference_scheme}
\end{figure}

During inference, object and wrist poses, as well as joint angles, are continuously sensed to provide feedback. After the policy prediction and decoding, $H$ actions are actuated from the predicted shifted sequence, where the horizon parameter must satisfy $H < n + N - L$. At the beginning of the demonstrations, the timestep buffer is padded to reach the input sequence length $L$. The algorithm \ref{alg:inference} and Fig. \ref{fig:inference_scheme} provide further clarification.

\begin{algorithm}[h!]
\caption{Behavior Cloning Inference Loop}\label{alg:inference}
\begin{algorithmic}
\State \textbf{define} H, L 
\State \textbf{sense} $s_{obj}, s_{r}$
\State timesteps  $\gets \{s_{r,1},...,s_{r,L}\}$  \Comment{Pad the sequence}
\While{not success}
\State $T \gets$ $|$timesteps$|$
\State $\tau_{input} \gets  $ timesteps[T-L:T]   \Comment{Take the last L steps}
\State $ z \gets \pi(\tau_{input},s_{obj}) $  \Comment{Predict Latent Space}
\State $ \tau_{output} \gets p_{\psi}(z) $ \Comment{Decode into actions}
\For{$k \gets 1$ ... $H$}  
\State \textbf{actuate} $\tau_{output,L-n+k}$
\State \textbf{sense} $s_{obj}, s_{r}$
\State timesteps $\gets$ timesteps $\cup$ $\{s_{r}\}$
\EndFor
\EndWhile
\end{algorithmic}
\end{algorithm}

\section{ANALYSIS}
\subsection{Simulation Results}
\begin{figure}
    \centering
    \includegraphics[width=0.49\textwidth]{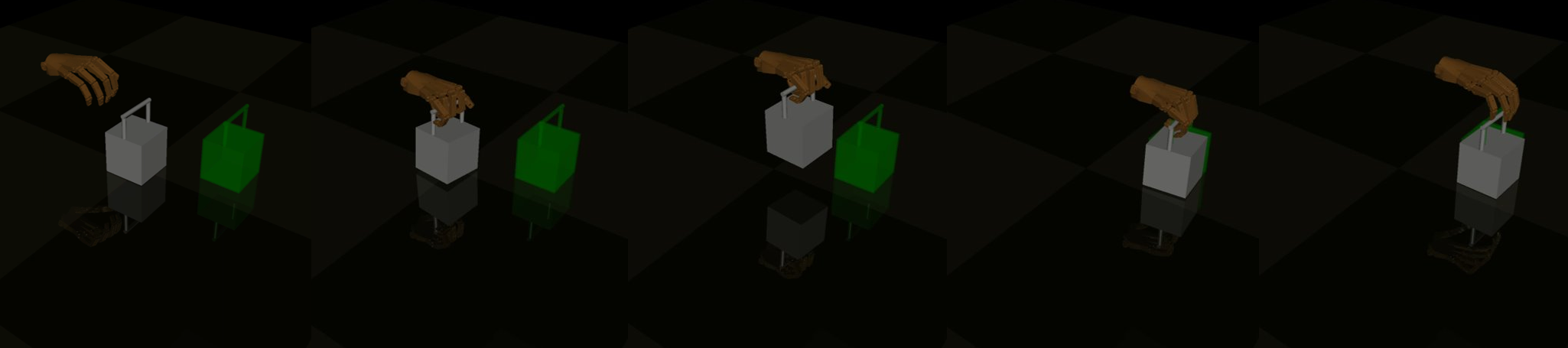}
    \caption{Relocation of a box with an handle, used for the simulation evaluation of the proposed work. The green shaded area represents the final goal used to compute the final error.}
    \label{fig:simulation_sequence}
\end{figure}

\begin{figure}
    \centering
    \includegraphics[width=0.49\textwidth]{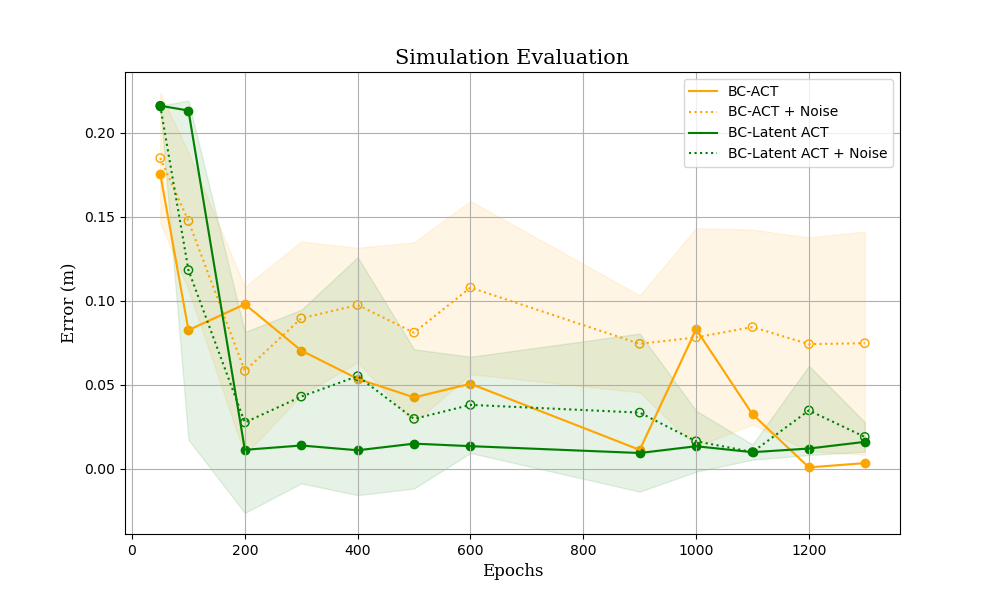}
    \caption{Position error of the relocated object during the training process. The policy was trained with 15 demonstrations. It can be noticed that our method (BC-Latent ACT) converges faster to accurate solutions. Moreover, the robustness is improved when external noise is introduced.}
    \label{fig:sim_experiment_num_demos}
\end{figure}

\begin{figure}
    \centering
    \includegraphics[width=0.49\textwidth]{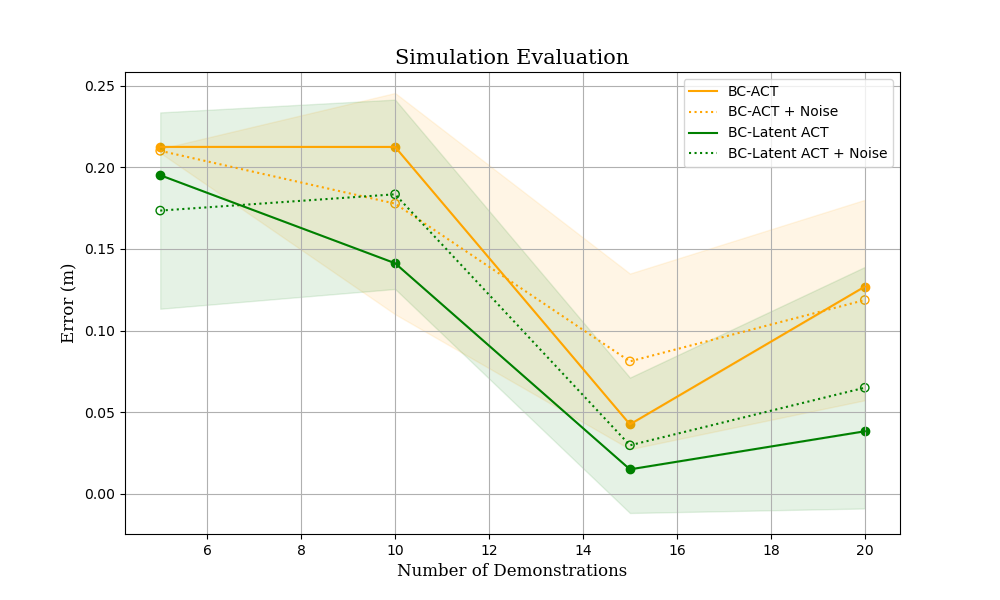}
    \caption{Position error of the relocated object with respect to the number of demonstrations used during training. The training time was fixed to 500 epochs. We can notice that our method performs better also in low data regimes.}
    \label{fig:simulation experiment_epochs}
\end{figure}

To analize the benefit of using pretrained latent representations in behavior cloning methods, we evaluated the accuracy of a pick-and-place task in the MuJoCo simulation environment \cite{mujoco}. The chosen object was a 10cm cube with a handle (fig \ref{fig:simulation_sequence}). In this task, the fingers must accurately be inserted into the handle and then the entire hand is lifted to pick up the object, allowing us to evaluate the coordination of the robot's joints. 
We recorded 20 demonstrations with the proposed pipeline, for a total of $\sim$ 15 min of data acquisition time. As comparison, we created a policy which has the same architecture but outputs robot actions directly, instead of latent representations. 
The metric on which the task was evaluated is the distance between the final position of the relocated object and the goal, computed as mean of the final position from the recorded demonstrations. The starting position of the hand was set to the average start position of the dataset.
During inference, object poses $s_{obj}$ and robot proprioception $s_{r}$ are continuously sensed from the simulation. We tested the trained policies also with artificial noise applied to the feedback inputs, to see how robust are the policies and how they adapt to inaccurate perception and proprioception. The applied noise was sampled from a Gaussian distribution with a standard deviation of 2 deg for the robot joints, 5 mm for the robot and object position, and 1 deg for the robot and object orientation.  
Results are shown in fig \ref{fig:simulation experiment_epochs} and \ref{fig:sim_experiment_num_demos}, and shows that the latent space significantly improves the resilience to noise. In particular, after 1200 epochs of training, the final error when noise is added is reduced by $\sim$ 83 \% This happens because by moving in the learned space of valid motion rather than the raw joint coordinates, the robot is less prone to sudden jumps, and the overall motion is more regular and smooth. From figure  \ref{fig:sim_experiment_num_demos} it can also be seen that our method converges faster to more accurate solutions, reducing training time by 75\% when 15 demonstrations are used.


\subsection{Real World Results}
We deployed the policies to the a real robotic dexterous system, to demonstrate successful transfer to real world performances. We used the hardware setup described in section \ref{hardware}. For tasks that relies on friction, silicon pads were added to the fingertips. 
The current hardware iteration of the Faive hand does not contain joint encoders and uses the tendon lengths to reconstruct the joint angles \cite{toshimitsu2023getting}, which is not very accurate when the hand is subject to external forces (e.g. contact with objects). Therefore, for real world experiments we did not use the proprioceptive feedback but instead fed back the commanded actions as robot states. On the other hand, the object states are acquired with visual tags, in the same way of the data acquisition pipeline. \\
At inference time, the hand joints commands are sent to the Faive hand, and the wrist poses are given to the IK of the Franka arm that computes the arm joint angles.

\begin{figure*}[h]
    \centering
    \includegraphics[width=\textwidth]{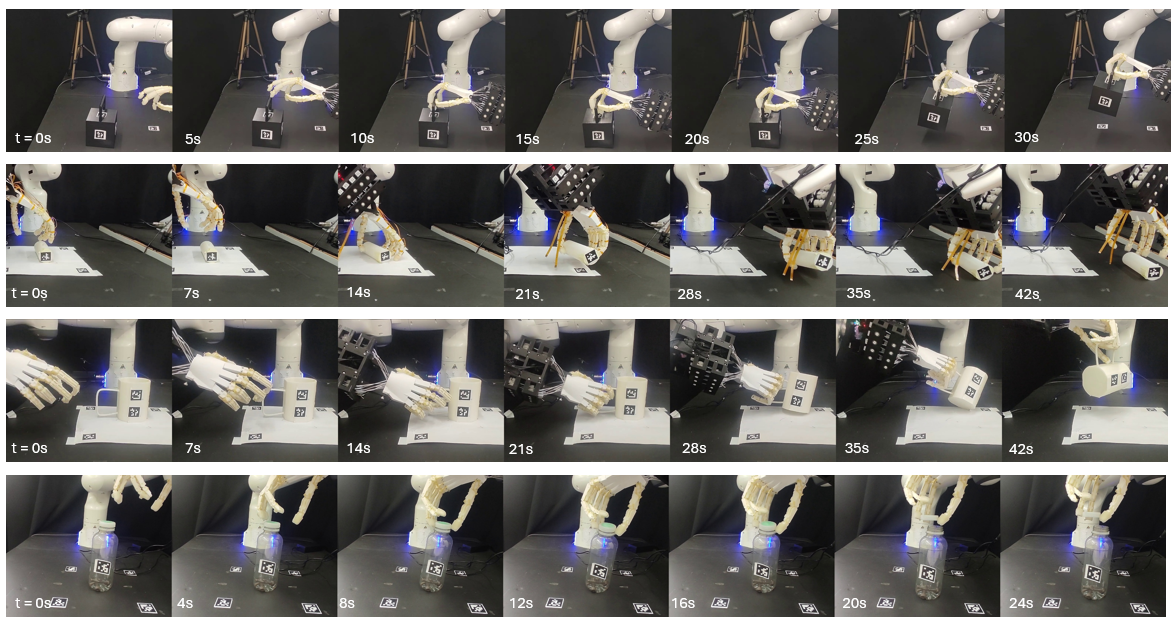}
    \caption{Examples of policies transferred to the real robot. From top to bottom: Handle grasping, cylinder relocation, mug grasping and pouring, cap unscrewing and picking. 
    The whole process of data collection, policy training and real-world deployment takes around 2 hours.}
    \label{fig:enter-label}
    
\end{figure*}
\section{DISCUSSION}
\subsection{Limitations and Future Work}
The presented work is subject to several limitations. Firstly, the input data heavily relies on the human-to-robot retargeter, which just takes into account fingertip positions. However for some tasks the manipulation is performed with other part of the hand. Indeed, if the retargeter is not able to effectively capture the interaction between the hand and the object, the policy will learn from inaccurate or wrong input data. This issue is less prevalent in teleoperation-based methods, where real-time visual feedback to the operator is available during the demonstration. 

Another important restricting factor is that the method exclusively uses position data as input and output, neglecting contact forces that are crucial to describe hand-object interactions. Consequently, tasks dependent on friction between the hand and the manipulated object pose a considerable challenge for successful replication.

\subsection{Conclusions}
In this work, we present a method to perform behavior cloning applied to a dexterous robotic hand that exploits pretrained latent representations. We show how to represent chunks of motion sequences that can be fed into a proposed VAE architecture so that the hand trajectory can be reconstructed, while also efficiently representing the motion in latent space.

We also develop a data acquisition process that relies just on human demonstrations without any teleoperation, enabling an efficient data collection process by non-experts and substantially speeding up the process with respect to teleoperation methods. We show that fingertip retargeting using IK can accurately transfer hand poses from the human hand data to the dexterous robot.

Experimental results, in which a transformer architecture was used as the policy, suggest superior performance compared to  BC methods without latent representations, demonstrating improved resilience to errors and noise in perception and proprioception. We have also applied the behavior cloning to various tasks requiring dexterity and precision on the real robot, such as picking and placing an object, or unscrewing a cap from a bottle.
These results suggest that latent space approaches can improve the performance of behavior cloning based on large behaviors models, since an efficient and consistent representation of actions can make results more robust.



\newpage
\section*{ACKNOWLEDGMENT}
The authors thank Gavin Barnabas Cangan for the support with Franka operations. Yasunori Toshimitsu is partially funded by the Takenaka Scholarship Foundation, the Max Planck ETH Center for Learning Systems, and the Swiss Government Excellence Scholarship.
This work was partially funded by the Amazon Research Awards.

\addtolength{\textheight}{-10cm}   


\begin{thebibliography}{}
\bibitem{faive} Faive Robotics https://www.faive-robotics.com/
\bibitem{openai2019learning}OpenAI, Andrychowicz, M., Baker, B., Chociej, M., Jozefowicz, R., McGrew, B., Pachocki, J., Petron, A., Plappert, M., Powell, G., Ray, A., Schneider, J., Sidor, S., Tobin, J., Welinder, P., Weng, L. \& Zaremba, W. Learning Dexterous In-Hand Manipulation.  (2019)
\bibitem{toshimitsu2023getting}Toshimitsu, Y., Forrai, B., Cangan, B., Steger, U., Knecht, M., Weirich, S. \& Katzschmann, R. Getting the Ball Rolling: Learning a Dexterous Policy for a Biomimetic Tendon-Driven Hand with Rolling Contact Joints.  (2023)
\bibitem{rajeswaran2018learning}Rajeswaran, A., Kumar, V., Gupta, A., Vezzani, G., Schulman, J., Todorov, E. \& Levine, S. Learning Complex Dexterous Manipulation with Deep Reinforcement Learning and Demonstrations.  (2018)
\bibitem{radosavovic2021stateonly}Radosavovic, I., Wang, X., Pinto, L. \& Malik, J. State-Only Imitation Learning for Dexterous Manipulation.  (2021)
\bibitem{ho2016generative}Ho, J. \& Ermon, S. Generative Adversarial Imitation Learning.  (2016)
\bibitem{shridhar2022perceiveractor}Shridhar, M., Manuelli, L. \& Fox, D. Perceiver-Actor: A Multi-Task Transformer for Robotic Manipulation.  (2022)
\bibitem{zhao2023learning}Zhao, T., Kumar, V., Levine, S. \& Finn, C. Learning Fine-Grained Bimanual Manipulation with Low-Cost Hardware.  (2023)
\bibitem{fu2024mobile}Fu, Z., Zhao, T. \& Finn, C. Mobile ALOHA: Learning Bimanual Mobile Manipulation with Low-Cost Whole-Body Teleoperation.  (2024)
\bibitem{brohan2023rt1}Brohan, A., Brown, N., Carbajal, J., Chebotar, Y., Dabis, J., Finn, C., Gopalakrishnan, K.,  \& Zitkovich, B. RT-1: Robotics Transformer for Real-World Control at Scale.  (2023)
\bibitem{brohan2023rt2}Brohan, A., Brown, N., Carbajal, J., Chebotar, Y., Chen, X., Choromanski, K., Ding, T., Driess, D., Dubey, A., Finn, C., Florence, P., Fu, C., Arenas, M., Gopalakrishnan, K., Han, K., Hausman, K., \& Zitkovich, B. RT-2: Vision-Language-Action Models Transfer Web Knowledge to Robotic Control.  (2023)
\bibitem{chi2023diffusion}Chi, C., Feng, S., Du, Y., Xu, Z., Cousineau, E., Burchfiel, B. \& Song, S. Diffusion Policy: Visuomotor Policy Learning via Action Diffusion.  (2023)
\bibitem{arunachalam2022dexterous}Arunachalam, S., Silwal, S., Evans, B. \& Pinto, L. Dexterous Imitation Made Easy: A Learning-Based Framework for Efficient Dexterous Manipulation.  (2022)
\bibitem{qin2023hand}Qin, Y., Su, H. \& Wang, X. From One Hand to Multiple Hands: Imitation Learning for Dexterous Manipulation from Single-Camera Teleoperation.  (2023)
\bibitem{wang2024cyberdemo}Wang, J., Qin, Y., Kuang, K., Korkmaz, Y., Gurumoorthy, A., Su, H. \& Wang, X. CyberDemo: Augmenting Simulated Human Demonstration for Real-World Dexterous Manipulation.  (2024)
\bibitem{qin2022dexmv}Qin, Y., Wu, Y., Liu, S., Jiang, H., Yang, R., Fu, Y. \& Wang, X. DexMV: Imitation Learning for Dexterous Manipulation from Human Videos.  (2022)
\bibitem{wang2024dexcap}Wang, C., Shi, H., Wang, W., Zhang, R., Fei-Fei, L. \& Liu, C. DexCap: Scalable and Portable Mocap Data Collection System for Dexterous Manipulation.  (2024)
\bibitem{chi2024universal}Chi, C., Xu, Z., Pan, C., Cousineau, E., Burchfiel, B., Feng, S., Tedrake, R. \& Song, S. Universal Manipulation Interface: In-The-Wild Robot Teaching Without In-The-Wild Robots.  (2024)
\bibitem{pertsch2021demonstrationguided}Pertsch, K., Lee, Y., Wu, Y. \& Lim, J. Demonstration-Guided Reinforcement Learning with Learned Skills.  (2021)
\bibitem{ajay2021opal}Ajay, A., Kumar, A., Agrawal, P., Levine, S. \& Nachum, O. OPAL: Offline Primitive Discovery for Accelerating Offline Reinforcement Learning.  (2021)
\bibitem{wang2023mimicplay}Wang, C., Fan, L., Sun, J., Zhang, R., Fei-Fei, L., Xu, D., Zhu, Y. \& Anandkumar, A. MimicPlay: Long-Horizon Imitation Learning by Watching Human Play.  (2023)
\bibitem{nasiriany2022learning}Nasiriany, S., Gao, T., Mandlekar, A. \& Zhu, Y. Learning and Retrieval from Prior Data for Skill-based Imitation Learning.  (2022)
\bibitem{ye2023learning}Ye, J., Wang, J., Huang, B., Qin, Y. \& Wang, X. Learning Continuous Grasping Function with a Dexterous Hand from Human Demonstrations.  (2023)
\bibitem{gehring2023leveraging}Gehring, J., Gopinath, D., Won, J., Krause, A., Synnaeve, G. \& Usunier, N. Leveraging Demonstrations with Latent Space Priors.  (2023)
\bibitem{rokoko} ROKOKO smart gloves https://www.rokoko.com/products/smartgloves
\bibitem{apriltags}Olson, E. AprilTag: A robust and flexible visual fiducial system 
\bibitem{zimmermann2017learning}Zimmermann, C. \& Brox, T. Learning to Estimate 3D Hand Pose from Single RGB Images.  (2017)
\bibitem{rong2020frankmocap}Rong, Y., Shiratori, T. \& Joo, H. FrankMocap: Fast Monocular 3D Hand and Body Motion Capture by Regression and Integration.  (2020)
\bibitem{liu2021semisupervised}Liu, S., Jiang, H., Xu, J., Liu, S. \& Wang, X. Semi-Supervised 3D Hand-Object Poses Estimation with Interactions in Time.  (2021)
\bibitem{rezaei2022trihornnet}Rezaei, M., Rastgoo, R. \& Athitsos, V. TriHorn-Net: A Model for Accurate Depth-Based 3D Hand Pose Estimation.  (2022)
\bibitem{MANO}Romero, J., Tzionas, D. \& Black, M. Embodied hands. {\em ACM Transactions On Graphics}. \textbf{36}, 1-17 (2017,11), https://doi.org/10.1145
\bibitem{wen2023bundlesdf}Wen, B., Tremblay, J., Blukis, V., Tyree, S., Muller, T., Evans, A., Fox, D., Kautz, J. \& Birchfield, S. BundleSDF: Neural 6-DoF Tracking and 3D Reconstruction of Unknown Objects.  (2023)
\bibitem{antotsiou2018taskoriented}Antotsiou, D., Garcia-Hernando, G. \& Kim, T. Task-Oriented Hand Motion Retargeting for Dexterous Manipulation Imitation.  (2018)
\bibitem{sivakumar2022robotic}Sivakumar, A., Shaw, K. \& Pathak, D. Robotic Telekinesis: Learning a Robotic Hand Imitator by Watching Humans on Youtube.  (2022)
\bibitem{handa2019dexpilot}Handa, A., Wyk, K., Yang, W., Liang, J., Chao, Y., Wan, Q., Birchfield, S., Ratliff, N. \& Fox, D. DexPilot: Vision Based Teleoperation of Dexterous Robotic Hand-Arm System.  (2019)
\bibitem{chao2021dexycb}Chao, Y., Yang, W., Xiang, Y., Molchanov, P., Handa, A., Tremblay, J., Narang, Y., Wyk, K., Iqbal, U., Birchfield, S., Kautz, J. \& Fox, D. DexYCB: A Benchmark for Capturing Hand Grasping of Objects.  (2021)
\bibitem{fan2023arctic}Fan, Z., Taheri, O., Tzionas, D., Kocabas, M., Kaufmann, M., Black, M. \& Hilliges, O. ARCTIC: A Dataset for Dexterous Bimanual Hand-Object Manipulation.  (2023)
\bibitem{Taheri_2020}Taheri, O., Ghorbani, N., Black, M. \& Tzionas, D. GRAB: A Dataset of Whole-Body Human Grasping of Objects. {\em Lecture Notes In Computer Science}. pp. 581-600 (2020), http://dx.doi.org/10.1007/978-3-030-58548-834
\bibitem{zhou2020continuity}Zhou, Y., Barnes, C., Lu, J., Yang, J. \& Li, H. On the Continuity of Rotation Representations in Neural Networks.  (2020)
\bibitem{mujoco}Todorov, E., Erez, T. \& Tassa, Y. MuJoCo: A physics engine for model-based control. {\em 2012 IEEE/RSJ International Conference On Intelligent Robots And Systems}. pp. 5026-5033 (2012)

\end{thebibliography}
\end{document}